\documentclass{article}

     \usepackage[preprint]{neurips_2020}

\usepackage[utf8]{inputenc} 
\usepackage[T1]{fontenc}    
\usepackage{hyperref}       
\usepackage{url}            
\usepackage{booktabs}       
\usepackage{amsfonts}       
\usepackage{nicefrac}       
\usepackage{microtype}      

\usepackage[ruled,vlined]{algorithm2e}
\usepackage{graphicx}
\usepackage{subfigure} 
\usepackage{color}
\usepackage{enumitem}
\usepackage{multirow}

\title{More Industry-friendly: Federated Learning with High Efficient Design}

\author{Dingwei Li\textsuperscript{1}\textsuperscript{2}\thanks{Work performed while doing an internship at Huawei Technologies Co., Ltd} \ \footnotemark[2] , Qinglong Chang\textsuperscript{1}\thanks{Contributted equally} , Lixue Pang\textsuperscript{1}, Yanfang Zhang\textsuperscript{1},Xudong Sun\textsuperscript{1}, Jikun Ding\textsuperscript{1}, Liang Zhang\textsuperscript{1}\\
  \textsuperscript{1}Huawei Technologies Co., Ltd\\
  \textsuperscript{2}Nanjing University, Nanjing 210023, China\\
  \texttt{\{lidingwei1,changqinglong,panglixue1,zhangyanfang2,}\\
  \texttt{sunxudong1,dingjikun1,zhangliang1\}@huawei.com}
}

\begin{document}

\maketitle

\begin{abstract}
Although many achievements have been made since Google threw out the paradigm of federated learning (FL), there still exists much room for researchers to optimize its efficiency. In this paper, we propose a high efficient FL method equipped with the double head design aiming for personalization optimization over non-IID dataset, and the gradual model sharing design for communication saving. Experimental results show that, our method has more stable accuracy performance and better communication efficient across various data distributions than other state of art methods (SOTAs), makes it more industry-friendly.
\end{abstract}

\section{\label{Introduction}Introduction}

Although FL is born for high efficient distributed model building, it’s not as efficient enough as we thought. Many existing methods may perform well for IID data setting, but non-ideal for non-IID data setting. In these methods,thier continuous global model sharing strategy, which makes them over focusing on global data distribution fitting but ignoring local data distribution. In FL, the methodology for this kind of problem is called personalization. To improve the personalization performance of FL, we bring the double head design in this paper. This idea is inspired from the multi-task learning, where multiple heads design  \citep{DBLP:journals/corr/LiH16e, DBLP:journals/corr/abs-1905-07835} is introduced to remember the model info from history missions. Similarly, we expect our double head design will achieve a good balance for model optimization between global and local data distribution.

Communication efficiency is another concerned topic in FL. Here we create the gradual sharing design to handle it. We get this idea from the analysis of model convergence rules. As we know, clients’ models share a similar convergence direction during their initial rounds of training. So it is not necessary to share model in this stage during this stage. In addition, we notice that once a frozen model restarts to share, it could recover to the normal performance effectively.  All this gives us confidence to try the gradual sharing design instead of always model sharing in FL to save communication cost.

We summarize three main contributions of our method as below.
%

\begin{itemize}[leftmargin=*]
\item We propose a double head design for FL, which makes our FL model benefit from the local and global data simultaneously. 

\item We introduce a gradual model sharing strategy for FL to save its communication cost greatly.

\item We split the data distribution setting over train and test data, which contributes to better evaluation for FL methods over various data distribution settings.  
\end{itemize}

The rest of our paper is organized as follow. In Sec.\ref{RelatedWork} we overview existing methods towards FL efficient problems. In Sec.\ref{OurDesign} we introduce our creative designs in detail. In Sec.\ref{Experiments} extensive experiments are made to evaluate our method. Finally, in Sec.\ref{Conclusion}, we summarize our method.

\section{\label{RelatedWork}Related Work}

Since the concept of FL was proposed by Google \citep{mcmahan2017communication}, personalization effectiveness and communication efficiency have always been concerned by researchers. 

\subsection{Personalization Effectiveness}

As mentioned before, the classic FL method has the drawback of handling non-IID data \citep{10.1145/3286490.3286559}. Roughly speaking, the principle idea for this problem is located at heterogeneous model parameter optimization, which is attempting to adapt a global model by fine-tuning, data-sharing or model-mixture. For the fine-tuning method \citep{wang2019federated, li2020lotteryfl}, a global model is given to each client in FL, then each cliect retrains the model with a new objective function based on their local data.. In the data-sharing method, some researchers \citep{DBLP:journals/corr/abs-1806-00582} try to upload a small part of local data to the server to tackle the non-IID problem, although it’s an obvious violation of data privacy. Other researchers attempt to build a generative model \citep{DBLP:journals/corr/abs-1811-11479} to cope with clients’ sample insufficient problem under the non-IID setting. As to the model-mixture method, researchers \citep{hanzely2020federated, deng2020adaptive, huang2020personalized} try to optimize the FL’s overall performance on global and local data distributions by ensembling models on server and clients. Other researchers attempt to achieve this by freezing data sharing on the last layer \citep{yang2020heterogeneity} of clients’ model. Currently, most of methods achieve better personalization performance at the price of more complicated computation cost and a sacrifice of performance on the IID data setting. 

\subsection{Communication Efficiency}

In the field of multi-task learning, some researchers aim to alleviate communication cost by compressing the communication content or reducing model update frequency \citep{tang2020communication}. Similar ideas continue in the FL area. At first, researchers \citep{DBLP:journals/corr/KonecnyMYRSB16, DBLP:journals/corr/Alistarh0TV16} propose to compress the model through a combination of quantization, subsampling, or encoding before sending it to the server. Then, other researchers \citep{DBLP:journals/corr/abs-1804-03235, DBLP:journals/corr/abs-1811-11479, yu2020salvaging} try to implement knowledge distillation \citep{hinton2015distilling} to save communication cost for the model aggregation. Recently, besides choosing sketched info to communicate over clients and server \citep{DBLP:journals/corr/abs-1903-04488}, more researchers start to optimize the communication efficient in FL by only synchronizing sub-model of the base model \citep{liang2020think, yang2020heterogeneity, li2020lotteryfl}.  

Several methods bear similar ideas with us. In method \citep{yang2020heterogeneity}, the last layer of the model is locked while training. As for method \citep{liang2020think}, the local model is divided into local and global part, the local part's model sharing is forbidden while training.  In our method, we take similar measures of constraining the model sharing across layers, while our specific freeze strategy is quite different from them. In addition, our method achieves better stable performance across various data distribution settings than those two methods. The method \citep{wang2020federated} also has a layer-wise model sharing strategy, which seems like our gradual sharing design.  However, it is optimized for the probabilistic model and can be seen as an extension for the bayesian nonparametric FL \citep{yurochkin2019bayesian} in CNN and LSTM setting. As a contrast, our method aims for a general model setting, and owns a completely different communication saving strategy.

\section{\label{OurDesign}Our Design}

\begin{figure*}[htbp]
  \subfigure{
		\begin{minipage}[t]{0.47\linewidth}
			\centering
  			\includegraphics[width=1\linewidth]{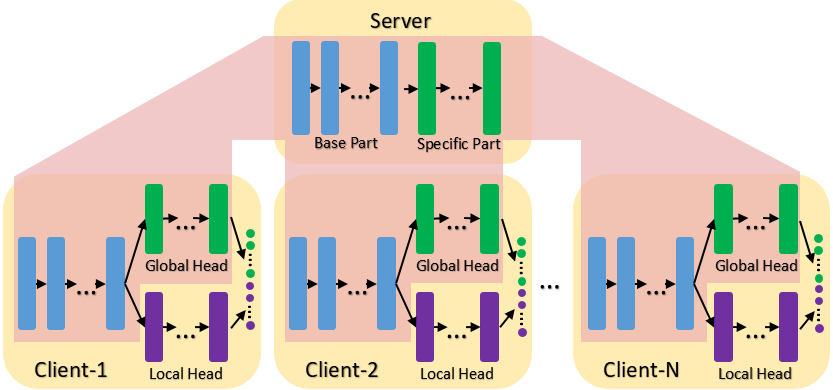}
  			\setcounter{figure}{0}\caption{\label{fig:DoubleHeadDesign}Double Head Design}
  		\end{minipage}
  }
  \quad
  \subfigure{
		\begin{minipage}[t]{0.45\linewidth}
			\centering
  			\includegraphics[width=1\linewidth]{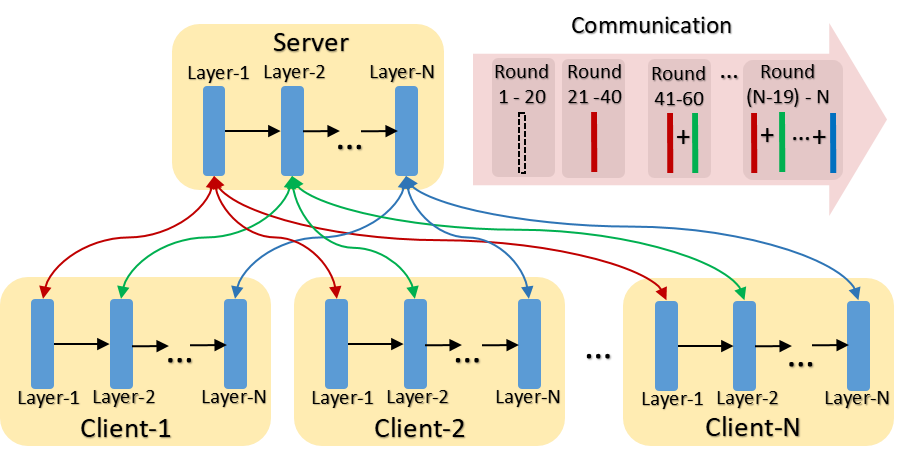}
  			\caption{\label{fig:GradualSharingDesign}Gradual Sharing Design}\setcounter{figure}{3}
  		\end{minipage}
  }
\end{figure*}

\subsection{Double Head}

As discussed before, our double head design is originated from the multiple head design in multi-task learning. These two heads are responsible for fitting the global and local data distribution respectively. For the sake of convenience, we take a sketch map of a classification task's network as an example to illustrate our theory in the follow-up, as shown in Figure \ref{fig:DoubleHeadDesign}. In our design, the network of the model could be divided into two main parts: the base part and the specific part.In the specific part, there're two tracks of network structure, each of which is consisted with several full connected layers and a softmax layer. These tracks are called the global and the local head respectively. Only the local head is forbidden model sharing while training, other parts of the model will always join model sharing. We formula the global and local heads’ output as

\begin{equation}
f_g(\overrightarrow{u})_i = \frac{e^{u_i}}{\sum_{j=1}^Ce^{u_j}}, \mbox{ for } i = 1, ..., C \mbox{ and } j = 1, ..., C
\end{equation}
\begin{equation}
f_l(\overrightarrow{v})_i = \frac{e^{v_i}}{\sum_{j=1}^Ce^{v_j}}, \mbox{ for } i = 1, ..., C \mbox{ and } j = 1, ..., C
\end{equation}

where the \(f_g\) and \(f_l\) are the output of a client model’s global and local head. \(\overrightarrow{u}\) and \(\overrightarrow{v}\) represent the penultimate output from the two heads. \(u_i\) and \(v_i\) mean the \(i\)-th element of vector \(\overrightarrow{u}\) and \(\overrightarrow{v}\). Parameter \(i\) and \(j\) are index of classes range from \(1\) to \(C\).

The final prediction result of the model is extracted from the output of these two heads’ softmax layers. We take the index of max value in the two output vectors as result, which is formulated as

\begin{equation}
{pred} = \mathop{\arg\max}_i(f_g(\overrightarrow{u})_i \oplus f_l(\overrightarrow{v})_i)
\end{equation}

where \(f_g\), \(f_l\), \(\overrightarrow{u}\) and \(\overrightarrow{v}\) bear the same meaning as in previous formulas; \(\oplus\) represents for the concatenation operation.

The complete pseudo-code of our double head design is given in Algorithm \ref{algorithmDHFL}, where \(K\) is clients' quantity in out FL system, \(T\) is the round number of model training, \(E\) is the number of local epochs, \(\eta\) represent the learning rate, \(w\) is the weight of server's model, \(wb, wg\) represent for the model weights of the base part and the global part, and \(w'\) is the weight of clients' model.

\begin{figure}
	\begin{minipage}{0.5\textwidth}
		\begin{algorithm}[H]
	\caption{\label{algorithmDHFL}Double Head Design}
	\SetKwProg{ServerExecutes}{Server executes}{:}{end}
	\SetKwProg{ClientUpdate}{ClientUpt}{:}{end}
	\SetKwProg{ClientInit}{ClientInit}{:}{end}
	\SetKwFunction{ClientUpdateFn}{ClientUpt}
	\SetKwFunction{ClientInitFn}{ClientInit}
	\SetKwFunction{concat}{concat}
	\SetKwFunction{split}{split}
	\SetKwFunction{valueFn}{value}
	
	\ServerExecutes{}{
		initialize \(w_0 \rightarrow\) \concat{\(wb_0, wg_0\)} \\
		\ForEach{client \(k = 1, ..., K\) in parallel}{
			\ClientInitFn{\(k, w_0\)} \\
		}
		\For{\(t = 1, 2, ..., T\)}{
			\(s_t \leftarrow (\mbox{random set of } m \mbox{ clients})\) \\
			\ForEach{client \(k \in s_t\)}{
				\(w_{t+1}^k \leftarrow\) \ClientUpdateFn{\(k, w_t\)} \\
			}
			\(w_{t+1} \leftarrow \sum_{k=1}^K\frac{n_k}{n}w_{t+1}^k\) \\
		}
	}
	\ClientUpdate{\((k, w_t)\)}{
		\(wb_t, wg_t \leftarrow\) \split{\(w_t\)} \\
		\({wb'}_{t-1}^k, {wg'}_{t-1}^k, {wl'}_{t-1}^k \newline \leftarrow\) \split{\({w'}_{t-1}^k\)} \\
		\({w'}_t^k \leftarrow\) \concat{\(wb_t, wg_t, {wl'}_{t-1}^k\)} \\
		\For{local epoch \(i = 1, 2, ..., E\)}{
			\({w'}_t^k \leftarrow {w'}_{t}^k - \eta \nabla l({w'}_t^k; D_k)\) \\
		}
		\({wb'}_{t}^k, {wg'}_{t}^k, {wl'}_{t}^k \leftarrow\) \split{\({w'}_{t}^k\)} \\
		\({w'}_t^{t+1} \leftarrow\) \concat{\({wb'}_{t}^k, {wg'}_{t}^k\)} \\
		\Return \(w_{t+1}^k\) to server \\
	}
	\ClientInit{\((k, w_0)\)}{
		\(wb_0, wg_0 \leftarrow\) \split{\(w_0\)} \\
		\({wb'}_0^k, {wg'}_0^k, {wl'}_0^k \leftarrow\) \valueFn{\(wb_0, wg_0, wl_0\)} \\
		\({w'}_0 \leftarrow\) \concat{\({wb'}_0^k, {wg'}_0^k, {wl'}_0^k\)} \\
	}
		\end{algorithm}
	\end{minipage}
	\quad
	\begin{minipage}{0.5\textwidth}
		\begin{algorithm}[H]
	\caption{\label{algorithmGSFL}Gradual Sharing Design}
	\SetKwProg{ServerExecutes}{Server executes}{:}{end}
	\SetKwProg{ClientUpdate}{ClientUpt}{:}{end}
	\SetKwProg{GetMask}{GetMask}{:}{end}
	\SetKwFunction{ClientUpdateFn}{ClientUpt}
	\SetKwFunction{GetMaskFn}{GetMask}
	\SetKwFunction{concat}{concat}
	
	\ServerExecutes{}{
		initialize \(w_0\) \\
		\ForEach{phase \(p = 1, 2, ..., P\)}{
			\(m_p \leftarrow\) \GetMaskFn{\(p\)} \\
			\For{\(t = 1, 2, ..., T\)}{
				\(s_t \leftarrow (\mbox{random set of } m \mbox{ clients})\) \\
				\ForEach{client \(k \in s_t\)}{
					\(ws_t \leftarrow m_p \otimes w_t\) \\
					\(ws_{t+1}^k \leftarrow\) \ClientUpdateFn{\(k, p, ws_t\)} \\
					\(w_{t+1}^k \leftarrow\) \concat{\newline\(ws_{t+1}^k, (1 - m_p) \otimes w_t\)} \\
				}
				\(w_{t+1} \leftarrow \sum_{k=1}^K\frac{n_k}{n}w_{t+1}^k\) \\
			}
		}
	}
	\ClientUpdate{\((k, p, ws_t)\)}{
		\(m_p \leftarrow\) \GetMaskFn{\(p\)} \\
		\({w'}_t^k \leftarrow\) \concat{\(ws_t, (1-m_p) \otimes {w'}_{t-1}^k\)} \\
		\For{local epoch \(i = 1, 2, ..., E\)}{
			\({w'}_t^k \leftarrow {w'}_t^k - \eta\nabla l({w'}_t^k; D_k)\) \\
		}
		\(ws_{t+1}^k \leftarrow m_p \otimes {w'}_t^k\) \\
		\Return \(ws_{t+1}^k\) to server \\
	}
	\GetMask{\((p)\)}{
		\(m \leftarrow (\mbox{initialize model size mask with } 0)\) \\
		\ForEach{layer \(l = 1, 2, ..., p\)}{
			\(m \leftarrow (\mbox{turn } l \mbox{ part of mask into } 1)\)
		}
		\Return \(m\)
	}
		\end{algorithm}
	\end{minipage}
\end{figure}

\subsection{Gradual Sharing}

Unlike other FL methods' full size model sharing, our method implements a new gradual sharing design, which is actually a model gradual sharing strategy for FL. This design helps us achieve more communication saving without hurting FL’s accuracy performance. As shown in Figure \ref{fig:GradualSharingDesign}, we will freeze clients sharing model with the server in initial training stages. Then, along with the training, clients could be allowed to share their model in layer-wise manner from shallower to deeper at certain frequency. Similarly, to better illustrate our idea, we give its pseudo-code in Algorithm \ref{algorithmGSFL}, in which \(K, E, \eta, w, w'\) are same to Algorithm \ref{algorithmDHFL}, and \(P\) indicates the current model sharing phase, which is calculated from the gradual sharing frequency, with value ranges from 1 to number of layers.

\section{\label{Experiments}Experiments}

\subsection{Experimental Detail}

\paragraph{Datasets}

Two challenge datasets are used here to evaluate our method: the FEMINIST dataset \citep{DBLP:journals/corr/abs-1812-01097} and the TCP Traffic Classification (hereinafter called the ‘TTC’) dataset.  FEMINIST is a public dataset of image classification, containing over 800k samples of 62 classes. Each sample (\(28 \times 28 \times 1\) pic) has its own writers and there are more than 3k writers in the dataset. Its data characteristics like: with plentiful enough classes; samples of the same class with various label types, are quite suitable for federated learning simulation. As for TTC dataset, it is a proprietary dataset from Huawei for traffic classification task. There are about 900k samples (\(10 \times 1\) vector) of 34 classes in it. It offers TCP traffic data which is captured and desensitized from two various data scenarios. This makes the evaluation on it more believable for industrial application.

According to industrial scenarios might met, we set three kinds of data distribution:
\begin{itemize}[leftmargin=*]
\item The \textbf{IID} setting: dataset distribution across clients is same. For FEMINIST, each client owns a dataset with the same amount of every writer’s samples. For TTC, each client is equally allocated data from two scenarios.

\item The \textbf{non-IID} setting: dataset distribution over clients is different, but each client’s data could cover the whole labels. For FEMINIST, clients are given whole label included data from different writers. For TTC, clients’ dataset are distributed from different data scenarios.

\item The \textbf{dispatch} setting: different clients own samples of uncrossed classes. This is a relative extreme non-IID setting for FL, but it is common for industrial scenario and discriminative enough for FL methods evaluation.
\end{itemize}

It's worth noting that, for most of current FL methods, the test data distribution will follow as the train data. But in real industrial scenarios, future data distribution is unknown, thus makes us to extend our test data into two basic modes, as follows.

\begin{itemize}[leftmargin=*]
\item The global mode: a new added test mode, where the distribution of test data will be close to the combination of all training data, and all clients model share the same test data. 

\item The local mode: the classic test mode, where the distribution of test data on each client will inherit their local train data, and therefore each client’s test data is different.
\end{itemize}

\paragraph{Implementations}

 To simplify compuation, we take a 5 layer network as the basic network, consisted of 2 convolution (conv) layers, 2 fully connected (fc) layers and a softmax layer in sequential. In our method, the first two conv + pooling layers are set as the base part and the rest layers are duplicated twice as the specific part. As a contrast, we take the Google’s FedAvg method \citep{mcmahan2017communication} as baseline, and also make a comparison with the HDAFL method \citep{yang2020heterogeneity}, who also declared itself better than FedAvg over non-IID data setting. All methods are implemented by Tensorflow v1.14 on Ubuntu 16.04. All models will be trained for over 400 rounds, and communication frequency is set at every 2 model iterations. In FEMINIST dataset, we simulate 10 clients, each has 10k samples. In TCC dataset, we have 2 clients, each owns around 200k samples.

\subsection{Double Head Effect}

\begin{figure}[htbp]
  \centering
  \subfigure[IID + global test]{
		\begin{minipage}[t]{0.24\linewidth}
  			\label{fig:DBHeadGIid}
			\centering
  			\includegraphics[width=1.1\linewidth]{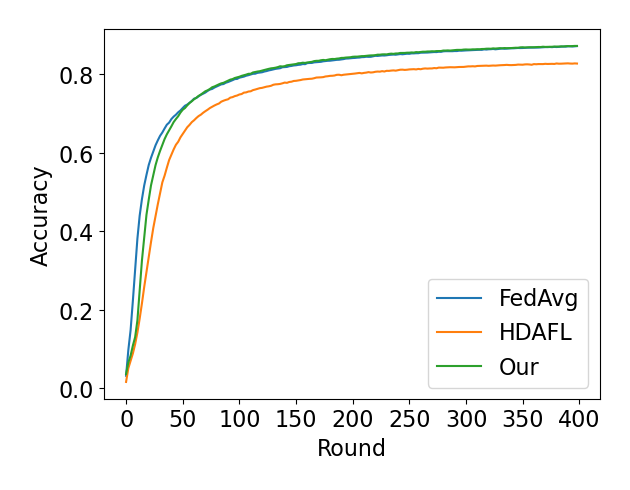}
  		\end{minipage}
  }
  \subfigure[non-IID + global test]{
		\begin{minipage}[t]{0.24\linewidth}
  			\label{fig:DBHeadGNonIid}
			\centering
  			\includegraphics[width=1.1\linewidth]{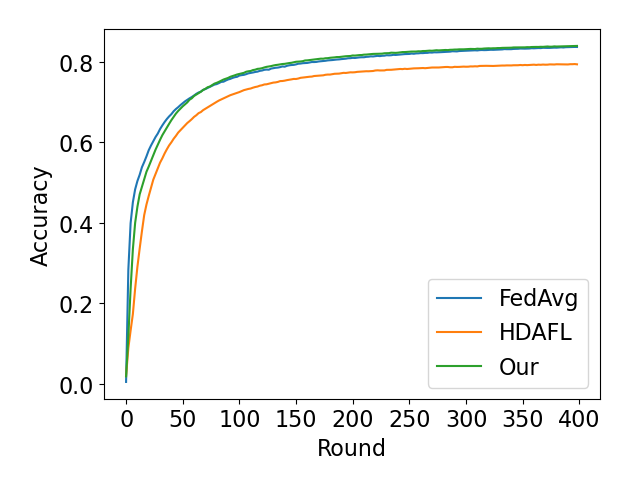}
  		\end{minipage}
  }
  \subfigure[dispatch + global test]{
		\begin{minipage}[t]{0.24\linewidth}
  			\label{fig:DBHeadGDispatch}
			\centering
  			\includegraphics[width=1.1\linewidth]{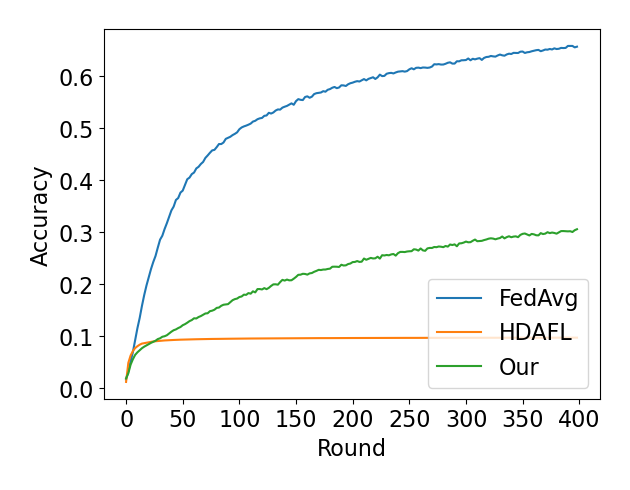}
  		\end{minipage}
  }
  \subfigure[IID + local test]{
		\begin{minipage}[t]{0.24\linewidth}
  			\label{fig:DBHeadLIid}
			\centering
  			\includegraphics[width=1.1\linewidth]{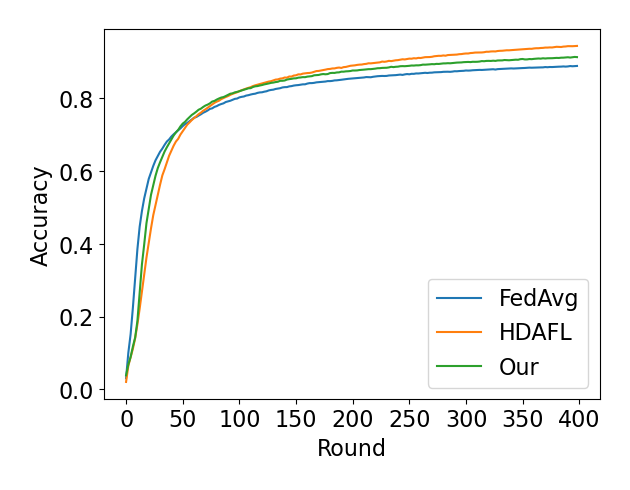}
  		\end{minipage}
  }
  \subfigure[non-IID + local test]{
		\begin{minipage}[t]{0.24\linewidth}
  			\label{fig:DBHeadLNoniid}
			\centering
  			\includegraphics[width=1.1\linewidth]{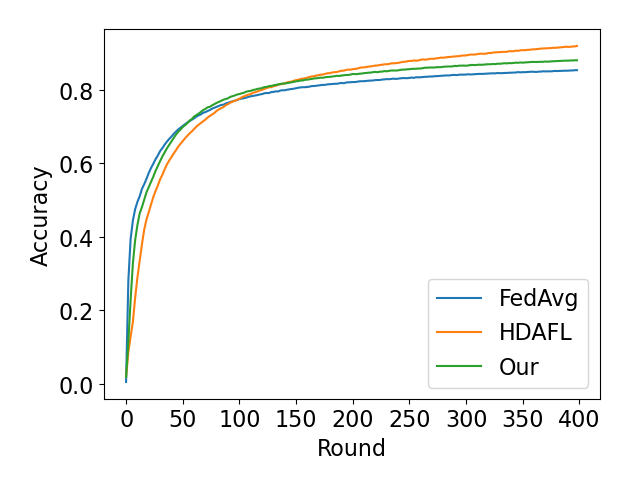}
  		\end{minipage}
  }
  \subfigure[dispatch + local test]{
		\begin{minipage}[t]{0.24\linewidth}
  			\label{fig:DBHeadLDispatch}
			\centering
  			\includegraphics[width=1.1\linewidth]{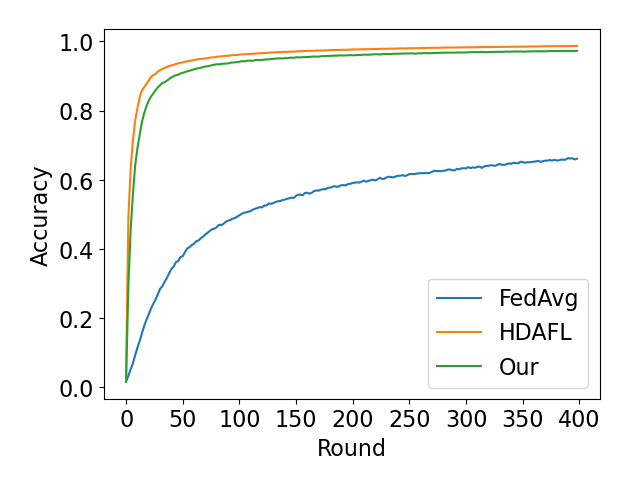}
  		\end{minipage}
  }
  \caption{\label{fig:DoubleHeadEffect}Compare of model accuracy performance over FEMNIST dataset}
\end{figure}

We firstly explore the advantages of double head design. As shown in Figure \ref{fig:DBHeadGIid} and \ref{fig:DBHeadGNonIid}, under the global test mode, with IID and non-IID data setting, our method has similar accuracy performance with FedAvg and performs better than HDAFL. This result can be explained by the global test data includes all clients' data distribution. Methods with more global model sharing power will gain more effective info while training. Likewise, method with less global sharing ability will achieve less in this setting.The result for IID and non IID data setting under local test mode is shown in Figure \ref{fig:DBHeadLIid} and \ref{fig:DBHeadLNoniid}. Our method performs a bit worse than HDAFL, but better than FedAvg. This might due to local test data mode setting, where methods with stronger fitting for local data distribution could perform better. In Figure \ref{fig:DBHeadGDispatch} and \ref{fig:DBHeadLDispatch}, the result of the dispatch data setting further supports above conclusions. In global test mode, FedAvg gets much better performance than others. While on local test mode, our method and HDAFL have similar performance and outperform FedAvg greatly. 

All this turns out our double head design enhance FL's fitting power for global and local data distribution simultaneously. Although it didn't get best accuracy result in all data settings, it performs more stable than other methods, which is more favorable in industry. After all, in industrial application, a better generalization performance is foremost, especially when future data distribution is unknown.

\subsection{\label{GradualSharingEffect}Gradual Sharing Effect}

We further explore the effectiveness of our gradual sharing design for FL.  First of all, we want to evaluate the impact of gradual sharing frequency. For FEMNIST dataset, according to our preliminary estimate, client’s model may reach convergence status after around 400 rounds of training. Accordingly, we set three types of sharing frequencies: release one layer every 10 rounds, every 20 rounds, and every 80 rounds respectively.

\begin{figure}[htbp]
  \centering
  \subfigure[IID + global test]{
		\begin{minipage}[t]{0.23\linewidth}
  			\label{fig:GSGIid}
			\centering
  			\includegraphics[width=1.1\linewidth]{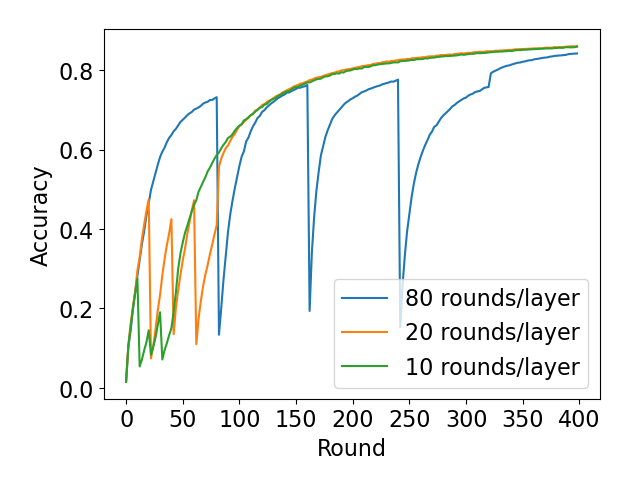}
  		\end{minipage}
  }
  \subfigure[non-IID + global test]{
		\begin{minipage}[t]{0.23\linewidth}
  			\label{fig:GSGNonIid}
			\centering
  			\includegraphics[width=1.1\linewidth]{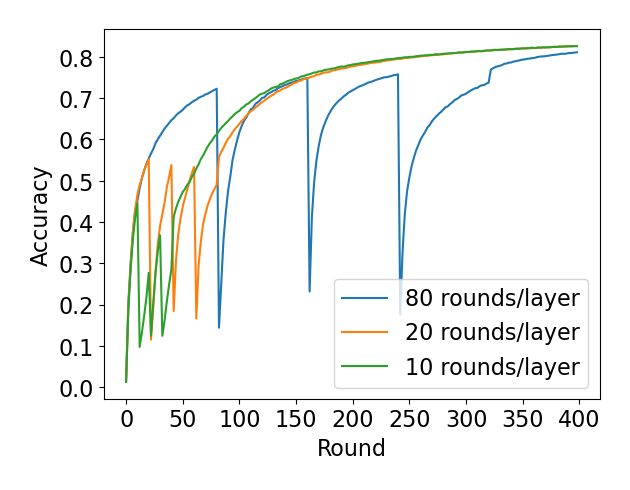}
  		\end{minipage}
  }
  \subfigure[dispatch + global test]{
		\begin{minipage}[t]{0.23\linewidth}
  			\label{fig:GSGDispatch}
			\centering
  			\includegraphics[width=1.1\linewidth]{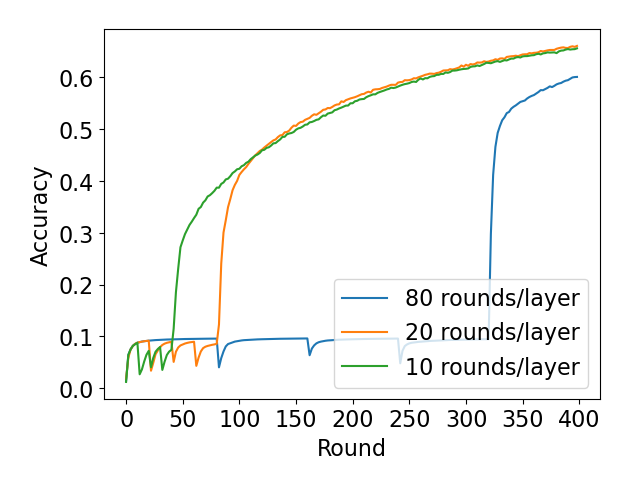}
  		\end{minipage}
  }
  \subfigure[IID + local test]{
		\begin{minipage}[t]{0.23\linewidth}
  			\label{fig:GSLIid}
			\centering
  			\includegraphics[width=1.1\linewidth]{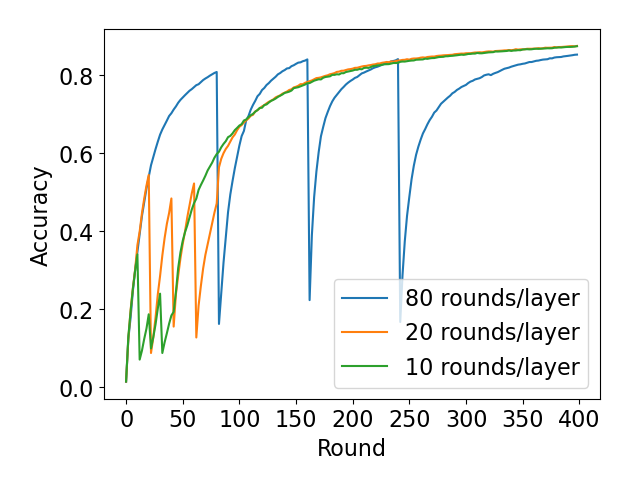}
  		\end{minipage}
  }
  \subfigure[non-IID + local test]{
		\begin{minipage}[t]{0.23\linewidth}
  			\label{fig:GSLNoniid}
			\centering
  			\includegraphics[width=1.1\linewidth]{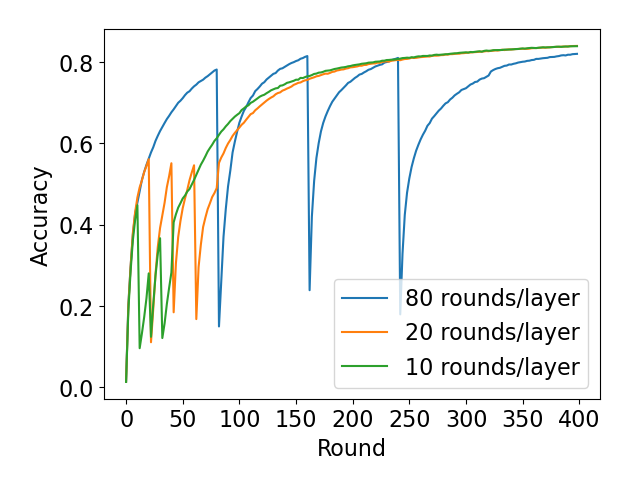}
  		\end{minipage}
  }
  \subfigure[dispatch + local test]{
		\begin{minipage}[t]{0.23\linewidth}
  			\label{fig:GSLDispatch}
			\centering
  			\includegraphics[width=1.1\linewidth]{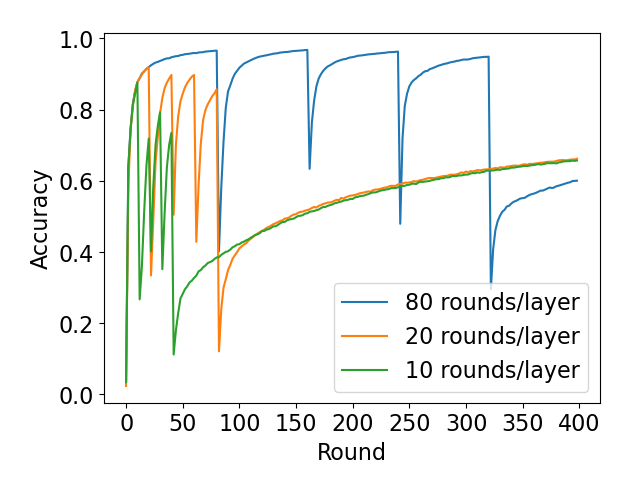}
  		\end{minipage}
  }
  \subfigure[FEMINST]{
		\begin{minipage}[t]{0.23\linewidth}
  			\label{fig:GSEffectTransmissionRate}
			\centering
  			\includegraphics[width=1.1\linewidth]{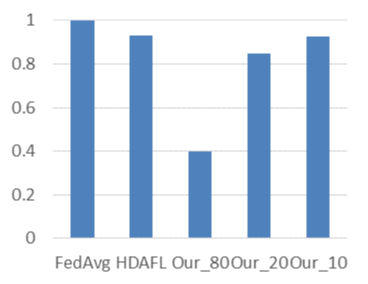}
  		\end{minipage}
  }
  \subfigure[TTC]{
		\begin{minipage}[t]{0.23\linewidth}
  			\label{fig:TTCGSEffectTransmissionRate}
			\centering
  			\includegraphics[width=1.1\linewidth]{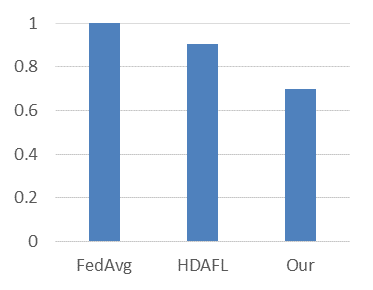}
  		\end{minipage}
  }
  \caption{\label{fig:GradualSharingEffect}Evaluation of Gradual Sharing Effectiveness: (a)-(f) are compare of convergence over various sharing frequencies; (g) and (h) are communication saving compare over two datasets}

\end{figure}

Results in Figure \ref{fig:GradualSharingEffect} has shown that, no matter in what kind of data setting or test mode, the gradual sharing applied model would resume to the normal accuracy performance quickly, although there would be a temporary decline once a layer sharing was relaxed. It’s worth noting that the frequency is an experimental parameter ranges from 0 to N (various according to dataset scenario). Once it's set too big, the model would be risky to resume to the normal accuracy within limited training rounds.


We also provide the comparison results for communication saving evaluation. In Figure \ref{fig:GSEffectTransmissionRate}, our method could achieve over 50\% communication saving than other two methods when gradual sharing frequency is set at 80 rounds/layer. This proves that our gradual sharing design would help FL system to save communication cost greatly without obvious accuracy performance decline.

\subsection{\label{ComprehensiveCompareInTTCDataset}Evalutaion on TTC dataset}

The TTC dataset is extracted from the telecom scenario, so an evaluation on this dataset could give a better reflect of our method’s performance in real industrial scenarios. The quantitative results in Table \ref{tab:QCOfModelAccPerfOverTTCDataset} show that our method's good accuracy performance across  different data settings. And the result in Figure \ref{fig:TTCGSEffectTransmissionRate} further proves that our method have an obvious advantage in communication saving. Besides, our method even performs better in TTC dataset than it is in FEMNIST dataset. This might due to its bigger model complexity induced by the double head design, which improves model’s ability to tackle a complex dataset like TTC.

\begin{table}
  \centering
  \caption{\label{tab:QCOfModelAccPerfOverTTCDataset}Quantitative compare of model accuracy performance over TTC dataset}
  \begin{tabular}{l|lll|lll}
    \toprule
    & \multicolumn{3}{c}{Global Test} & \multicolumn{3}{c}{Local Test} \\
    \cmidrule(r){2-7}
    Name     & IID & non-IID & dispatch & IID & non-IID & dispatch \\
    \midrule
    FedAvg & 0.8375 & 0.8341 & \textcolor{red}{0.7994} & 0.8503 & 0.8385 & 0.7374 \\
    HDAFL & 0.8370 & 0.8021 & 0.4420 & 0.8498 & 0.8583 & 0.8792 \\
    Our\_DH & \textcolor{red}{0.8544} & \textcolor{red}{0.8512} & 0.7495 & \textcolor{red}{0.8680} & \textcolor{red}{0.8722} & 0.8723 \\
    Our\_DH+GS & 0.8430 & 0.8444 & 0.7472 & 0.8562 & 0.8704 & \textcolor{red}{0.8818} \\
    \bottomrule
  \end{tabular}
\end{table}

\section{\label{Conclusion}Conclusion}

Improving the personalization effectiveness and the communication efficient are always the research focus in FL. In this paper, we propose the double head design and the gradual sharing design to tackle these challenges. Our experimental results show that the double head design effectively enhance FL's accuracy performance under the non-IID data setting. And the gradual sharing design could save communication cost hugely without impeding model convergence. Although our method didn't get the best accuracy result over all data settings, it achieves a more stable performance across various test data settings compared to other SOTAs. This helps it to be more industry attractive.

\bibliographystyle{plainnat}
\bibliography{neurips}

\end{document}